\documentclass[conference]{IEEEtran}
\IEEEoverridecommandlockouts
\usepackage{tikzpeople}
\usepackage{cite}
\usepackage{amsmath,amssymb,amsfonts}
\usepackage{listings}
\usepackage{algorithmic}
\usepackage{float}
\usepackage{tikz}
\usepackage{graphicx}
\usepackage{textcomp}
\usepackage{xcolor}
\usepackage{url}
\usepackage[final]{pdfpages}
\usepackage{hyperref}

\usetikzlibrary{shapes.callouts}
\usetikzlibrary{positioning,arrows,calc}

\tikzset{modal/.style={>=stealth',shorten >=1pt,shorten <=1pt,auto,node distance=1.5cm,semithick},world/.style={circle,draw,minimum size=0.5cm,fill=gray!15},point/.style={circle,draw,inner sep=0.5mm,fill=black},reflexive above/.style={->,loop,looseness=7,in=120,out=60},reflexive below/.style={->,loop,looseness=7,in=240,out=300},reflexive left/.style={->,loop,looseness=7,in=150,out=210},reflexive right/.style={->,loop,looseness=7,in=30,out=330}}

\def\BibTeX{{\rm B\kern-.05em{\sc i\kern-.025em b}\kern-.08em
    T\kern-.1667em\lower.7ex\hbox{E}\kern-.125emX}}
    
    \definecolor{lightGray}{gray}{0.9}
\begin{document}

\newcommand{\nb}[2]{
    \fbox{\bfseries\sffamily\scriptsize#1}
    {\sf\small\textcolor{red}{\textit{#2}}}
}
\newcommand\ag[1]{\nb{AG}{#1}}
\newcommand\bi[1]{\bi{LT}{#1}}
\newtheorem{puzzle}{Puzzle}

\title{Natural language understanding for logical games}

\author{\IEEEauthorblockN{Adrian Groza and Cristian Nițu}
\IEEEauthorblockA{\textit{Department of Computer Science,} \\
\textit{Technical University of Cluj-Napoca}\\
Cluj-Napoca, Romania \\
\tt Adria.Groza@cs.utcluj.ro, Nitu.Co.Cristian@utcluj.didatec.ro}
}

\maketitle

\begin{abstract}

We developed a system able to automatically solve logical puzzles in natural language. 
Our solution is composed by a parser and an inference module. 
The parser translates the text into first order logic (FOL), while the MACE4 model finder is used to compute the models of the given FOL theory.
We also empower our software agent with the capability  to provide Yes/No answers to natural language questions related to each puzzle. 
Moreover, in line with Explainalbe Artificial Intelligence (XAI), the agent can back its answer, providing a graphical representation of the proof. 
The advantage of using reasoning for Natural Language Understanding (NLU) instead of Machine learning is that the user can obtain an explanation of the reasoning chain. 
We illustrate how the system performs on various types of natural language puzzles, including 382 knights and knaves puzzles.
These features together with the overall performance rate of 80.89\% makes the proposed solution an improvement upon similar solvers for natural language understanding in the puzzles domain.
\end{abstract}


\section{Introduction}
\begin{figure}
\begin{center}
\begin{tikzpicture}[scale=1,every node/.style={transform shape},roundnode/.style={circle, draw=green!60, fill=green!5, very thick, minimum size=7mm},]
\node[name=b,shape=dave,female, shirt=orange, mirrored, minimum size=1cm,xshift=-6cm] {Rachel};
\node[name=a,shape=bob,shirt=red,mirrored, minimum size=1cm,xshift=-3cm] {Ross};
\node[name=c,shape=alice,mirrored,minimum size=1cm,xshift=0cm] {Monica};
\node[ellipse callout, draw, align=center, above left= 5pt and 0pt of a.north west, callout absolute pointer={(a.mouth)}, font=\tiny] {I am in a break!};
\node[ellipse callout, draw, align=center, above left= 5pt and 0pt of b.north west, callout absolute pointer={(b.mouth)}, font=\tiny] {I am not married};
\node[ellipse callout, draw, align=center, above left= 5pt and 0pt of c.north west, callout absolute pointer={(c.mouth)}, font=\tiny] {I am married};
\end{tikzpicture}
\end{center}
\caption{A warm-up puzzle: Is a married person looking at an unmarried one?}
    \label{fig:friends}
\end{figure}
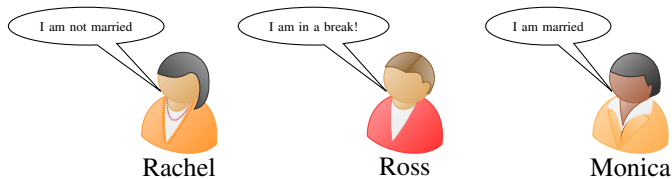

Here is a puzzle for you: \textit{There are three friends staying on the couch in Central Perk: Rachel, Ross, and Monica.
Monica is looking at Ross. Ross is looking at Rachel. Monica is married; Rachel is not.
Is a married person looking at an unmarried person?}~\cite{groza2021} (Figure~\ref{fig:friends}).
The answer may seem difficult since we do not know if Ross is married or not. 
A fan of the ``Friends'' series would know that even Ross has some difficulties to figure out if he is married or not.

Yet, such puzzles are easy to solve for theorem provers, given the proper formalisation. 
Take for instance the given facts in first order logic (FOL): 
\begin{center}
$~~married(Monica)$ \hfill $looking(Monica,Ross)$\\
$\neg married(Rachel)$ \hfill  $looking(Ross,Rachel)$
\end{center}
The theorem to prove:
\begin{equation}
 \exists x\ \exists y\ (married(x) \wedge \neg married(y) \wedge looking(x,y))
\end{equation}
is easily demonstrated by resolution-based theorem provers. 
The  challenge remains how to translate the text of the puzzle from natural language to first order logic.

Building a system that can understand the natural language as well as a human is considered a "too hard" AI problem~\cite{yampolskiy2013turing}, since an exact understanding of all clues is necessary to figure out the solution. 
Barrière views language understanding as a transformation of text into a deeper representation, one on which reasoning can occur~\cite{barriere2016natural}. 
In this line, we argue here that solving logical puzzles by machines is a powerful and relevant test to illustrate natural language understanding. 

For a software agent, solving such puzzles raises two main challenges: 
i) translating the text from natural language to some logical representation; and ii) reasoning on that formalisation to infer the solution. 
 Also, in most of the cases, background knowledge is required to solve the puzzle (e.g. the married relation is symmetric  $married(x,y) \leftrightarrow married(y,x)$, and functional $married(x,y) \wedge married(x,y) \rightarrow y=z$). 
 What distinguishes puzzles from classical tasks in natural language understanding is their nice property: each piece of information provided in the text is required. Hence, a good puzzle provides no text which is not useful. 
 Differently from tasks from the knowledge representation domain in which there is a lot of knowledge (e.g. large ontologies, big data) and little reasoning, here puzzles require knowledge and reasoning in similar proportions.

We present a solver for logical puzzles in natural language. 
We rely on grammar rules and lexical resources such as Wordnet~\cite{miller1995wordnet}. 
The text is automatically translated into FOL using the NLTK~\cite{bird2006nltk} and then given to the theorem prover Prover9 and model finder Mace4~\cite{mccune2005prover9}.

\section{System overview}
\begin{figure}
\centering
\includegraphics[width=.6\textwidth]{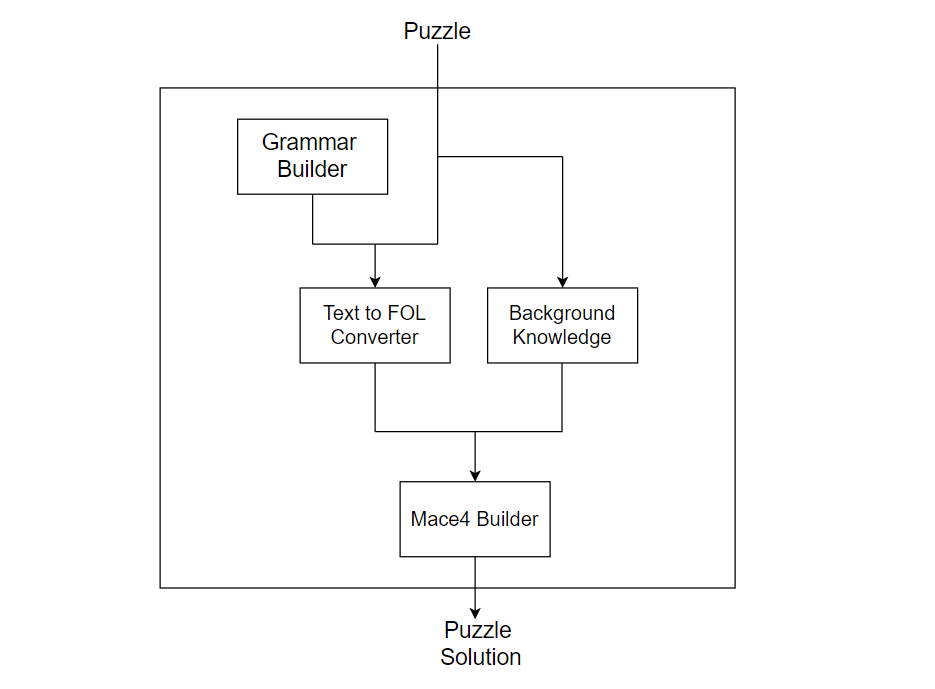}
    \caption{The four modules of the puzzle solver}
    \label{fig:architecture}
\end{figure}

The input of this system consists of the puzzle to be solved, in natural language. 
The solver\footnote{The system is available at - \href{ https://github.com/NituCristian/AutomaticWordPuzzlesSolverSystem}{Link}} follows four steps (Figure~\ref{fig:architecture}). 
First, we have to build the context-free grammar which contains the lexicon and the grammar. 
Second, the text is translated into FOL. 
Third, we add the background knowledge for the given domain (i.e. knights and knaves puzzles). 
Fourth, a model builder (in our case, Mace4) is called to compute the solutions. 

\subsection{Building the grammar}
Consider the following small puzzle:
\textit{Diana is taller than Maria.
Ana is taller than Diana.
Who is the shortest?}
The grammar for such puzzles contains both lexical and grammar rules: 

\textit{Lexical rules} hold the linguistic categories for each puzzle (Listing~\ref{lst:lexical}). 
A custom encoding is defined for every category (e.g. N for noun, VT for transitive verbs, IV for intransitive verbs). 
Every word has to belong to such a category and can have one or more \textit{features}~\cite{wagner2010steven}. 
Two of the most important features used by the solver are named \texttt{SEM}, which is used to specify the semantics of the word and \texttt{NUM}, which can be used to define the number of a noun (i.e. which can have a value like $sg$ for singular or $pl$ for plural). 
Thus, an invalid sentence as "These dog barks" cannot be parsed. 
To assess the correctness of sentences, we also included features  like \texttt{PERS} for distinguish the person, or \texttt{TENSE} for verb tenses. 

The most important feature is the one for defining the semantics of the words. 
The values of the \texttt{SEM} feature are computed based on $\lambda$-calculus. 
For example, the expression $\backslash x.(walk(x)\ \&\ chew\_gum(x))$ represents the entities that both walk and chew gum. If we apply this expression to a person named Gerald, we obtain the expression $\textbackslash x.(walk(x) \& chew\_gum(x)) (gerald)$, which is equivalent to the expression $walk(gerald) \& chew\_gum(gerald)$, meaning that Gerald walks and chews gum. 
The lambda operator is essential in connecting different words.

\begin{lstlisting}[caption=Lexical rules,linewidth=\columnwidth,breaklines=true,firstline=22,lastline=35,label=lst:lexical]
PropN[-LOC,NUM=sg,SEM=<\P.P(kevin)>] -> 'Kevin'
PropN[-LOC,NUM=sg,SEM=<\P.P(diana)>] -> 'Diana'
PropN[-LOC,NUM=sg,SEM=<\P.P(maria)>] -> 'Maria'

Det[NUM=sg,SEM=<\P Q.exists x.((P(x) & Q(x)) & all y.(P(y) -> (x = y)))>] -> 'the' | 'The' 

InterPron[-LOC, NUM=sg, SEM=<\P.P(x)>] -> 'who' | 'Who'

Aux[+COP,NUM=sg,SEM=<\P x.P(x)>,tns=pres] -> 'is'
Aux[+COP,NUM=pl,SEM=<\P x.P(x)>,tns=pres] -> 'are'

Adj[GRD=rel, SEM=<\X x.X(\y.(taller(x,y)))>] -> 'taller'
Adj[GRD=abs, SEM=<\x.(shortest(x))>] -> 'shortest'
P[+than] -> 'than'
\end{lstlisting}

 \textit{Grammar rules} define, in a recursive way, how words from different parts of speech are connected ). 
 The rules in Listing~\ref{lst:lexical} were manually crafted for the application domain.
 These sequences of words are grouped into \textit{chunks}. 
 An example of such a chunk is the noun phrase chunk, which can be exemplified by expresions like "the man" or "the red ball". 
 Unlike the words in lexical rules, which can be considered terminals, the chunks are nonterminals. 
 A grammar is context-free if the left side of its rules are only nonterminals, whereas the right-hand side can be a combination of terminals and nonterminals.
 
Important for consistency and generalization is the usage of variables as values for features. 
Thus, using a wildcard operator, we can define a rule like 
\begin{equation}
S[NUM=?n] \rightarrow (A[NUM=?n]\ B[NUM=?n])     
\end{equation}

which says that the feature $NUM$ of both $A$ and $B$ must have the same value, so we can control the correctness of a sentence.
Note that the context-free grammar in Listing~\ref{lst:grammar}  can parse sentences such as "Kevin is taller than Diana" or "Maria is the tallest", but not "Maria are the tallest".
 
\begin{lstlisting}[caption=Grammar rules,linewidth=\columnwidth,breaklines=true,firstline=5,lastline=18,label=lst:grammar]
S[SEM = <?subj(?vp)>] -> NP[NUM=?n,SEM=?subj] VP[NUM=?n,SEM=?vp]

NP[LOC=?l,NUM=?n,SEM=?np] -> PropN[LOC=?l,NUM=?n,SEM=?np] | InterPron[LOC=?l,NUM=?n,SEM=?np]

VP[NUM=?n,SEM=<?v(?prd)>] -> AuxP[+COP,NUM=?n,SEM=?v] Pred[SEM=?prd]

Pred[SEM=?prd] -> AdjP[GRD=?rel, SEM=?prd] | AdjP[GRD=?abs, SEM=?prd]

AdjP[GRD=?rel, SEM=<?adj(?pp)>] -> Adj[GRD=?rel, SEM=?adj] PP[+than,SEM=?pp]
AdjP[GRD=abs, SEM=?adj] -> Det[NUM=sg] Adj[GRD=abs, SEM=?adj] 

PP[+than, SEM=?np] -> P[+than] NP[SEM=?np]

AuxP[COP=?c,NUM=?n,SEM=?aux] -> Aux[COP=?c,NUM=?n,SEM=?aux]
\end{lstlisting}

\subsection{Translating to First-Order Logic}
We start the translation mechanism with the \textit{coreference resolution}, followed by \textit{parsing} based on the grammars previously built.

\textit{Coreference resolution} aims to make a connection between different expressions from the text and the corresponding entities (e.g. the pronoun \textit{he} with the corresponding person that the pronoun refers to). 
Failing to achieve coreference resolution leads to failure in solving the puzzle, as some information would be omitted or interpreted in a wrong way. 
For this task we rely  on the \textit{Neuralcoref} module based on the \textit{spaCy} parser~\cite{vasiliev2020natural}.

Next,  we obtain a tree representation for every sentence in the updated puzzle. 
The representation of every sentence in FOL is obtained from the previously defined \texttt{SEM} feature, which is stored in the parse trees. 
The lexicon is automatically extended with all the synonyms offered by the Wordnet lexical database, so that 
a word without definition in the grammar file but with a synonym defined in the grammar file
can be successfully parsed.
The "unseen" word will take all the attributes from the synonym found in the grammar file, so there is no difference between the parsing result of a sentence containing unseen words and the one containing only predefined words. 

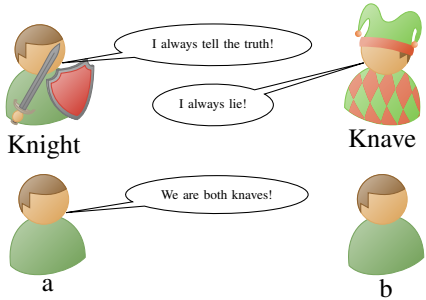
\begin{figure}
\centering
\begin{tikzpicture}[scale=1,every node/.style={transform shape},roundnode/.style={circle, draw=green!60, fill=green!5, very thick, minimum size=7mm},]
\node[name=knight,shape=person,shield,sword,minimum size=1cm,xshift=-2.25cm] {Knight};
\node[name=knave,shape=jester,minimum size=1cm, mirrored,xshift=2.25cm] {Knave};
\node[ellipse callout, draw, yshift= .4cm,  callout absolute pointer={(knight.mouth)}, font=\tiny] {I always tell the truth!};
\node[ellipse callout, draw, yshift=-.4cm, callout absolute pointer={(knave.mouth)}, font=\tiny] {I always lie!};
\end{tikzpicture}\vspace{0.5cm}

\begin{tikzpicture}[scale=1,every node/.style={transform shape},roundnode/.style={circle, draw=green!60, fill=green!5, very thick, minimum size=7mm},]
\node[name=a,shape=person,minimum size=1cm,xshift=-2.25cm] {a};
\node[name=b,shape=person,minimum size=1cm,xshift=2.25cm] {b};
\node[ellipse callout, draw, yshift= .4cm,  callout absolute pointer={(a.mouth)}, font=\tiny] {We are both knaves!};
\end{tikzpicture}

\caption{Example of puzzle with knights and knaves}
\label{fig:knights}
\end{figure}

As an example, let the following puzzle:
\begin{puzzle}\label{puz:knights}
\textit{On the island where each inhabitant is either a knave or a knight , knights always tell the truth while knaves always lie.
You meet two inhabitants: Marge and Homer. 
Marge says that Homer and she are both knights or both knaves.
Homer claims that Marge and he are the same. 
Can you determine who is a knight and who is a knave?} (Figure~\ref{fig:knights}).
\end{puzzle}

\begin{figure}
\includegraphics[scale=0.38]{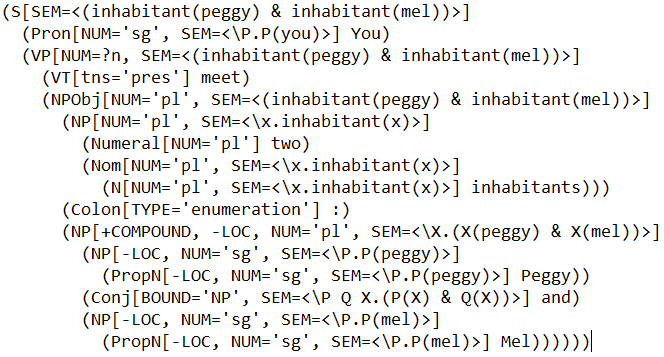}
    \caption{Parsing tree for "You meet two inhabitants:..."}
    \label{fig:tree}
\end{figure}

Figure~\ref{fig:tree} shows the parse tree for the first sentence. Note that there are some words whose semantics are not considered when the representation in FOL has to be obtained, like the pronoun "You" or the transitive verb "meet". 
By using the grammar rules the first sentence is translated into FOL  as three clauses:
\begin{eqnarray}
\forall x\ inhabitant(x) \rightarrow knave(x) \vee knight(x)\\
\forall x\ knight(x) \rightarrow truth(x) \\
\forall x\ knave(x) \rightarrow lie(x)
\end{eqnarray}
The sentence \textit{Marge says that Homer and Marge are both knights or both knaves} is formalised with:
\begin{eqnarray}
    say(marge) \leftrightarrow (knight(homer) \wedge knight(marge)) \nonumber \\
    \vee (knave(homer) \wedge knave(marge)) 
\end{eqnarray}
The clue \textit{Homer claims that Marge and Homer are the same} is formalised with:
\begin{equation}
  (claim(homer) \leftrightarrow same(homer,marge))  
\end{equation}
One of the system features, in contrast to other existing solvers, is the ability to distinguish, for example, between the person names from the puzzle. 
Thus, the user does not need to specify, by himself, who are the named entities in the puzzle he wants to solve. 
Recognising the named entities was achieved by the spaCy library, using a trained existing pipeline in order to get a label for the discovered named entities by making a prediction. 
The task in our example is to retrieve the existing persons in the puzzle. 
Since the unique name assumption does not hold by default in FOL, we explicitly state the distinctions between names (e.g. $alice \neq  sam$).

\subsection{Adding background knowledge}
Although for a human agent the task of figuring out the additional knowledge needed to reach a solution is an easy one, 
for the software agent it is challenging~\cite{groza2015information}. 


For each word in the puzzle, we extract from the WordNet the set of synonyms. 
Then, we check if any synonym from the set coexists in the specified text. 
If so, we formalise in FOL the equivalence (e.g. $\forall x\ (say(x) \leftrightarrow tell(x))$). 
When extracting the synonyms we considered the part of speech also. 
For example, the word "state" can be both a noun (e.g country) or a verb (as a synonym for "say"). 
Both of these two words are valid synonyms, but in different contexts. 
Therefore, we determine the POS in the current context and consider just the synonyms which have the same tag. 
To avoid building some assumptions which refer to the same words, but in different forms (like $\forall x\ (say(x) \leftrightarrow tell(x))$ and $\forall x\ (says(x) \leftrightarrow tell(x))$), lemmatization is applied. 

There exists, however, some contextual meaning that depends on the puzzle domain. 
For instance, in the puzzles of knights and knaves, the sentence "Alice and Sam are the same", requires the proper formalisation 
for the predicate "same": 
\begin{eqnarray}
\forall x \forall y\ same(x, y) \leftrightarrow (knight(x) \wedge knight(y)) \vee \\
(knave(x) \wedge knave(y))    
\end{eqnarray}
The domain knowledge for the knight and knaves puzzles states also that there are only knights and knaves:
\begin{eqnarray}
\forall x\ knave(x) \vee knight (x)  \\
\forall x\ knave(x) \rightarrow \neg knight (x)
\end{eqnarray}
The above tiny background knowledge for the knights and knaves domain was also manually crafted.

\subsection{Computing the solution}
\begin{figure}
\centering
\includegraphics[width=.3\textwidth]{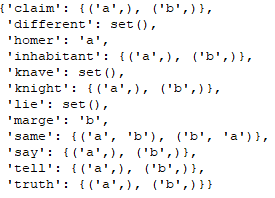}
\vspace*{0.4cm}
\begin{tikzpicture}[scale=1,every node/.style={transform shape},roundnode/.style={circle, draw=green!60, fill=green!5, very thick, minimum size=7mm},]
\node[name=a,shape=alice,shield,sword,minimum size=1cm,xshift=-3.25cm] {Marge};
\node[name=b,shape=person,shield,sword,minimum size=1cm,xshift=4.25cm] {Homer};
\node[ellipse callout, draw, yshift= .4cm,  callout absolute pointer={(a.mouth)}, font=\tiny] {Homer  and  I  are  both  knights  or  both knaves};
\node[ellipse callout, draw, yshift=-.4cm, callout absolute pointer={(b.mouth)}, font=\tiny] {Marge  and I are  the same};
\end{tikzpicture}

    \caption{Computing models with MACE4: both Homer (i.e. $a$) and Marge (i.e. $b$) are knights}
    \label{fig:solution}
\end{figure}
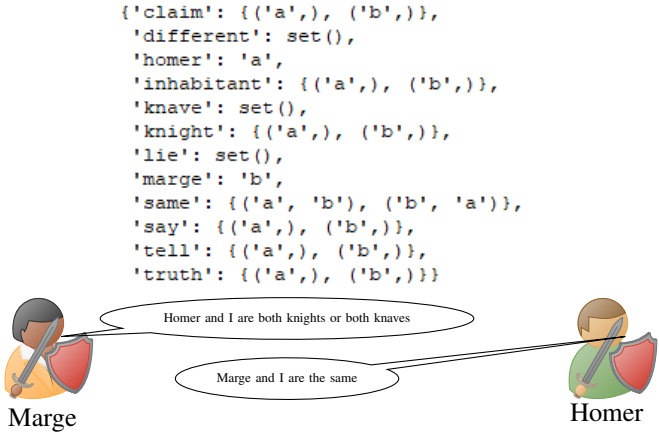

In the final step, the Mace4 model builder computes the solution. 
Mace4 uses the FOL representation of the puzzle to compute a model for the given FOL theory. 
For example, Mace4's solution for the Puzzle~\ref{puz:knights} appears in Figure~\ref{fig:solution}. 

Using a model finder to answer the puzzle is not necessarily
an obvious choice since the constraints could have more than one
model. However, fa good puzzle should have sufficient constraints to
ensure that the answer to the question does not depend on the choice
of model. 

\subsection{Question answering and explainability}
We empower our solver, with two interesting features: 
i) question answering (QA) - the capacity to answer various natural language questions from the puzzle, and 
ii) explainability (XAI)- the capability to provide graphical proofs for each answer provided. 
For these features we used Prover9 to provide Yes/No answers and graphical representation of the proofs for each answer. 

For exemplifying the question answering capability consider the following puzzle:
\begin{puzzle}
\textit{On the island where each inhabitant is either a knave or a knight, knights always tell the truth while knaves always lie. 
You meet two inhabitants: Sue and Alice. Sue claims that Alice is a knave. 
Alice says that she and Sue are knights.}
\end{puzzle}

First, the solver uses named entity resolution to identify the two persons Sue and Alice. 
Hence the domain size is set to two elements. 
Then the correference resolution figures out that the pronoun "she" refers to Alice in the sentence: "Alice says that she and Sue are knights." 
The Wordnet is used to identify that "claims and  "says" are synonyms, thus adding this piece of knowledge in the FOL theory.
The FOL theory built using the grammar file, the domain size, and the background knowledge is given to MACE4, which finds valid instantiations for each element in the theory. 
The user can ask questions like:
\begin{center}
\begin{tabular}{c}
    \textit{ Is Sue a knight?}  \\
     \textit{Does Alice lie?} \\
     \textit{Is Alice a knave?}\\
     \textit{Are Alice and Sue different?}\\
     \textit{Are Alice and Sue the same?}\\
     \textit{Are Alice and Sue both knights?}
\end{tabular}
\end{center}
The same grammar file is used to parse the queries and to translate them as theorems in First Order Logic. 
The theorem is given to Prover9 which provides a YES/No answer. 



\begin{figure}
\centering
\includegraphics[width=.48\textwidth]{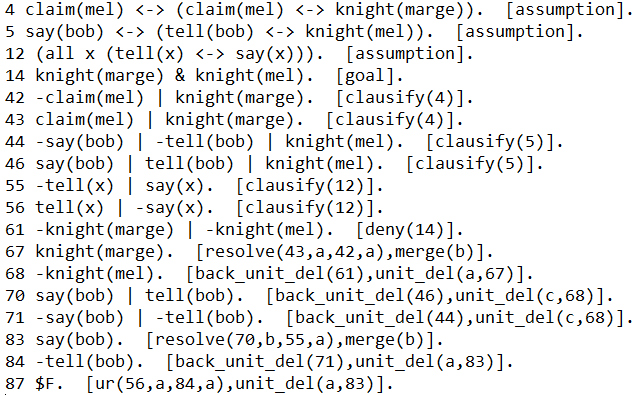}
    \caption{Proof for the computed solution}
    \label{fig:answer}
\end{figure}

Figure~\ref{fig:answer} shows the proof for the theorem  "Marge and Mel are knights". 
There are assumptions from text (line 4) or from synonymy between words (line 12), while the goal appears in line 14. 
Some clauses represent disjunctions of literals deduced from assumptions (line 42). 
Based on resolution, new clauses inferred (e.g. line 68), until the empty clause is deduced which signals that the theorem is proved.

For each Yes/No computed answer, the system outputs a graphical representation of the proof. 
To illustrate this feature, consider the following puzzle.
\begin{puzzle}
\textit{On the island of knights and knaves, knights always tell the truth, while knaves always lie. 
You are approached by two people. 
The first one says: ``We are both knaves''. }
\end{puzzle}
Consider also the compound query: \textit{Is the first inhabitant a knave and the second one a knight?}
The corresponding FOL theory appears in Listing~\ref{lst:kk0p}.

First, the inhabitants  are either knights or knaves:
\begin{equation}
 \forall x\ (inhabitant(x) \rightarrow knight(x) \vee knave(x)) 
\end{equation}
Second, one cannot be a knight and a knave in the same time: 
\begin{equation}
 \forall x\ ((knight(x) \rightarrow \neg knave(x)) \wedge (knave(x) \rightarrow \neg knight(x))).
\end{equation}
Third, a message $m(x)$ said by a knight $x$ is always true: $knight(x) \rightarrow  m(x)$.
Fourth, a message $m(x)$ said by a knave $x$ is always false: $knave(x) \rightarrow  \neg m(x)$.
These pieces of knowledge represent background knowledge for all knights and knaves puzzles 
Next, we formalise the current puzzle.
We learn that there are two inhabitants, let's say $a$ and $b$: $inhabitant(a) \wedge inhabitant(b)$. 
We need a domain size of two individuals.
The message of inhabitant $a$ is: $m(a) \leftrightarrow  knave(a) \wedge knave(b)$. 

\begin{lstlisting}[caption=A FOL theory used to provide Yes/No answers,linewidth=\columnwidth,breaklines=true,label=lst:kk0p]
formulas(assumptions).
  all x (inhabitant(x) -> knight(x) | knave(x)). 
  all x ((knight(x) -> -knave(x)) & (knave(x) -> -knight(x))).
  knight(x) -> m(x).              knave(x)  -> -m(x).
  
  inhabitant(a) & inhabitant(b).
  m(a) <->  knave(a) & knave(b). 
end_of_list.

formulas(goals).
  knave(a) & knight(b).
end_of_list.
\end{lstlisting}

\begin{figure*}
\begin{center}
\begin{tikzpicture}[scale=1,every node/.style={transform shape},roundnode/.style={circle, draw=green!60, fill=green!5, very thick, minimum size=7mm},]
\node[name=a,shape=jester,minimum size=1cm,xshift=-2.25cm] {a};
\node[name=b,shape=person,shield,sword, minimum size=1cm,xshift=2.25cm] {b};
\node[ellipse callout, draw, yshift= .4cm,  callout absolute pointer={(a.mouth)}, font=\tiny] {We are both knaves!};
\end{tikzpicture}

\includegraphics[angle =-90,width=\textwidth]{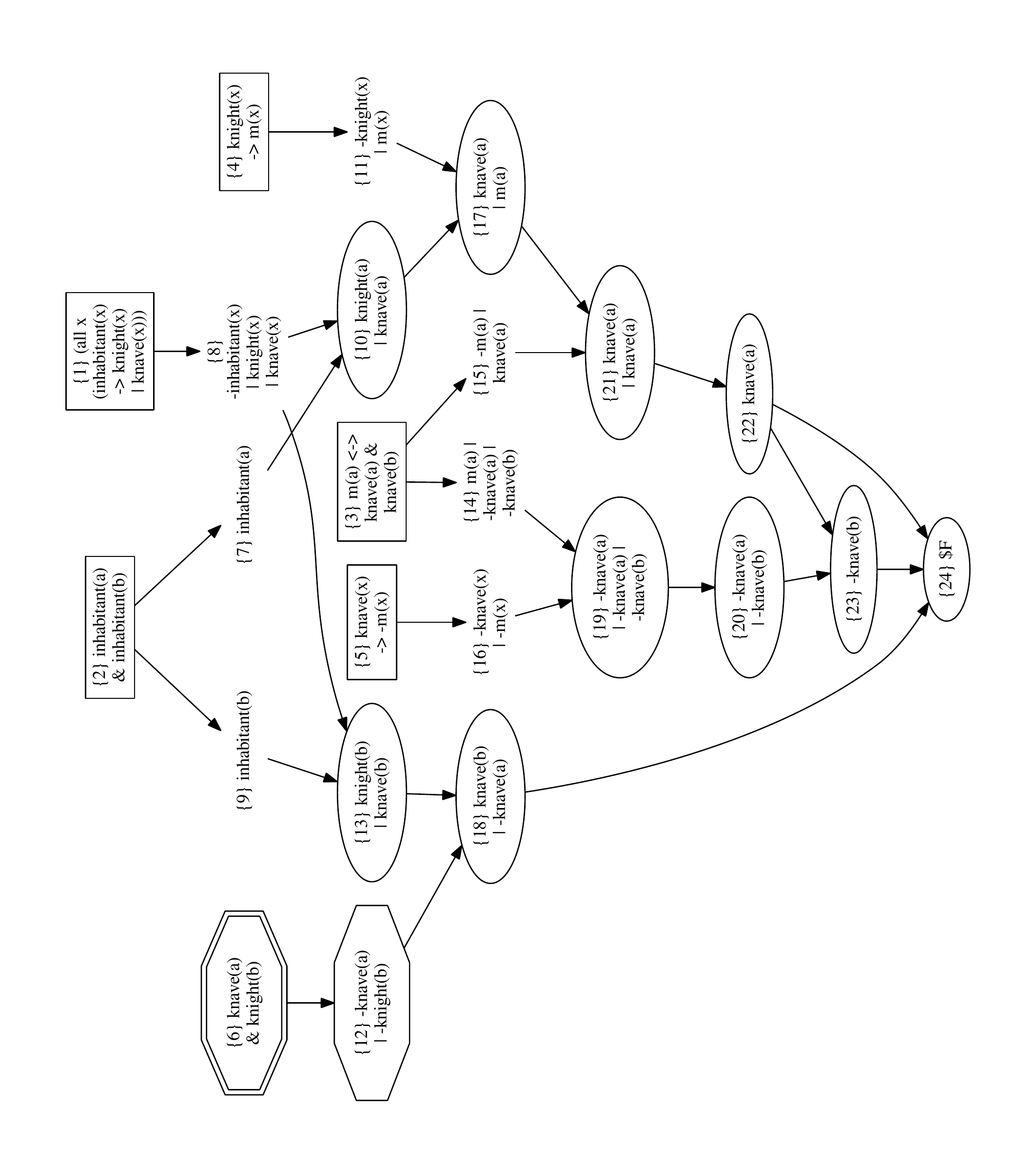}
\caption{Proof for clause \{6\}: $a$ is a knave and $b$ is a knight based on clause \{3\}: ``We are both knaves''\label{fig:kk0p}}
\end{center}
\end{figure*}

Given the code in Listing~\ref{lst:kk0p}, Prover9 finds the proof in Figure~\ref{fig:kk0p} for the goal $knave(a) \wedge knight(b)$.
Here, the prover finds a contradiction based on three pieces of knowledge. 
First, the negated conjecture \{12\} is equivalent to \{18\}: $knave(b) \vee \neg knave(a)$. 
Second, the agent deduces \{17\}: $a$ is a knave or the message is true. 
Further, based on \{15\}, the system deduces that $a$ is a knave (clause \{22\}).
Third, the prover infers \{20\} based on the input clause \{5\} (knaves are liars) and the given message \{3\}. 
Using resolution between \{20\} and \{22\}, $b$ is not a knave. 
By combining these three pieces of knowledge (\{18\}, \{22\}, \{23\}) through hyper-resolution, the software agent signals a contradiction. 
Hence, the theorem \{6\} is proved.

\section{Running experiments}
We evaluate the solver on a dataset of 382 puzzles with knights and knaves\footnote{Puzzles are taken from the website \\ https://philosophy.hku.hk/think/logic/knights.php}. 
The complexity of these puzzles is variable, starting with puzzles with two individuals (such as Puzzle~\ref{puz:knights}), continuing with puzzles with more text and persons such as Puzzle~\ref{puz:knightsDifficult}. As a measure of complexity, the puzzles can be classified based on the number of inhabitants: there are 50 puzzles each with two, three, four, five, six, seven and eight inhabitants, and 32 puzzles having nine inhabitants. 

\begin{puzzle}\label{puz:knightsDifficult}
\textit{"On the island where each inhabitant is either a knave or a knight, knights always tell the truth while knaves always lie.
You meet nine inhabitants: Carl, Betty, Ted, Dave, Marge, Alice, Rex, Bob and Sally. 
Carl claims that Ted would tell you that Alice is a knave. 
Betty says that at least one of the following is true: that Ted is a knave or that Dave is a knave. 
Ted says that Rex could say that Betty is a knave. Dave claims that Sally and Alice are different. 
Marge says that Rex is a knave.
Alice claims that Carl is a knave or Sally is a knight. 
Rex says that he knows that he is a knight and that Bob is a knave. 
Bob says that he and Rex are both knights or both knaves. 
Sally tells you that Bob is a knave. 
Can you determine who is a knight and who is a knave?"}
\end{puzzle}

We start by formalising the context-free grammar file by analysing 43 puzzles of different complexities, including 20 puzzles with two inhabitants, 5 puzzles with 3 inhabitants, and 3 puzzles each with four, five, six, seven, eight and nine inhabitants. 
The remaining 339 unseen puzzles were used to test the grammar.

\begin{table}
\centering
\caption{System performance on knights and knaves puzzles set}
\label{tab:tableResult}
\begin{tabular}{ |p{3.3cm}|p{1.2cm}|p{1.2cm}|p{1.4cm}| }
 \hline
 &Testing puzzzles &Solved puzzles &Performance\\
 \hline
 Grammar   &339    &331   &97.64\% \\
 \hline
 Named Entity Recognition    &382    &342 &89.52\% \\
 \hline
 Coreference Resolution   &382    &352 &92.14\% \\
 \hline
\end{tabular}
  
\end{table}

In our experiments, first we aimed to check how many of the unseen 339 puzzles can be automatically solved. Second, we aimed to identify the impediments for those puzzles that couldn't be solved. 
Hence, we analysed three points of failure: 
i) the incapacity of the grammar to parse the text, 
ii) the failure to identify all the persons from the puzzles, and 
iii) the failure of coreference resolution step.
Table~\ref{tab:tableResult} bears out the experimetal results against these three coordinates. 
In line 1, since 43 puzzles were used to build the grammar, just the rest of 339 puzzles are tested. 
The performance of the Named Entity Recognition (line 2) and Coreference Resolution (line 3) were tested against the entire set of 382 puzzles.
The conclusions for each metric are:

\begin{enumerate}
  \item \textit{Grammar}: The grammar manually built by analysing 43 puzzles manages to parse 97.64\% of the remaining 331 puzzles. 
  The 8 unsolved puzzles need more grammar rules to be covered 
  \item \textit{Named Entity Recognition}:  The implementation offered by spaCy encountered some problems in finding out all of the person names from our puzzles. For instance, there were some puzzles where a name like "Peggy" or "Bozo" were not recognized. 
  In some puzzles, although all the person names were found, the trained pipeline used made some other mistakes (e.g. it identified "Ted and Bob" as a single person instead to "Ted" and "Bob" as different persons). 
  \item \textit{Coreference Resolution}: Two problems were noticed: i) there were some pronouns which could not be bound to any person name; and ii) the pronoun was not replaced with the name of the person it should. Both cases led to the incapacity of the system to solve the puzzle further. However, the task of coreference resolution had a performance score of 92.15\%.
\end{enumerate}

We also analysed how the performance is affected by the puzzle's complexity. 
Here, we measure complexity in terms of number of persons appearing within the text (see Table~\ref{tab:persons}).
\begin{table}
\centering
\caption{Performance based on puzzles complexity (i.e. number of inhabitants)}
\label{tab:persons}
\begin{tabular}{ |p{3.3cm}|p{1.2cm}|p{1.2cm}|p{1.4cm}| }
 \hline
 Number of inhabitants &Puzzles & Solved puzzles & Performance\\ \hline
 2 & 50 & 44 & 88\% \\ \hline
 3 & 50 & 43 & 86\% \\ \hline
 4 & 50 & 42 & 84\% \\ \hline
 5 & 50 & 41 & 82\% \\ \hline
 6 & 50 & 39 & 78\% \\ \hline
 7 & 50 & 39 & 78\% \\ \hline
 8 & 50 & 38 & 76\% \\ \hline
 9 & 32 & 23 & 71.87\% \\ \hline
 \end{tabular}
  
\end{table}

\begin{figure*}
\centering
\includegraphics[width=\textwidth]{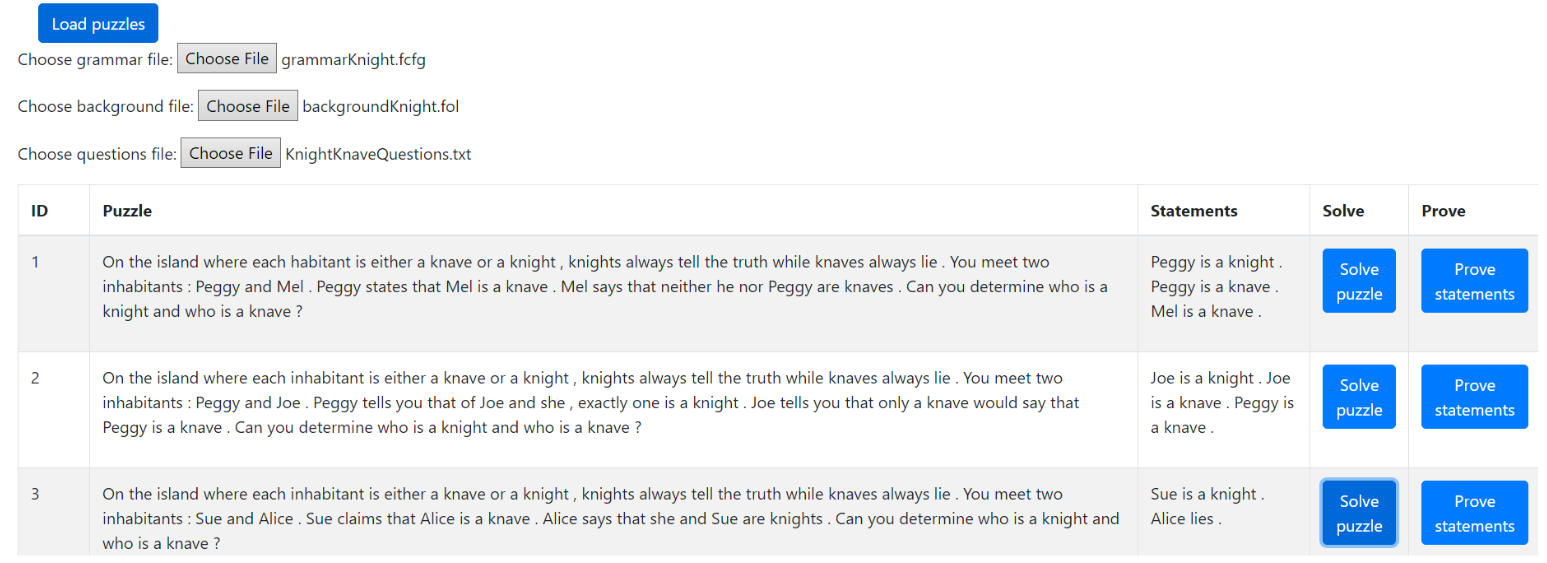}
    \caption{Running system}
    \label{fig:solver}
\end{figure*}

We focused here on knight and knaves puzzles. 
However, the system is extensible: to solve puzzles from a different domain, one has to provide the corresponding grammar file (see Figure~\ref{fig:solver}). 
The public version of our system provides small grammars for other types of puzzles. 

Taking into account that there were five puzzles where neither the Named Entity Recognition, nor the Coreference Resolution managed to be applied successfully, in total, a number of 73 puzzles out of 382 could not be solved. 
Hence, on the proposed set of puzzles, the overall performance of our solver is 80.89\%.

\section{Discussion and Related work}

Solving logical word puzzles is considered a challenging task. 
Lev et al. have also proposed a solution based on grammar rules, FOL, and model builders~\cite{lev2004solving}. 
Lev et al. have focused multiple-choice puzzles, so the inferences only have to find out the correct answer, not to discover it. 
Milicevic et al. have tackled the task by using the ink Grammar general-purpose English parser, a semantic translator, and an automated logical analyzer~\cite{milicevic2012puzzler}. 
 The Alloy language is used as a formal representation of the puzzle and and its corresponding constraints solver computes the solution. 
 The solver is designed around the Zebra puzzles and tested on a dataset of 68 puzzles.
Here, the user must rephrase the ambiguous phrases, which is a disadvantage of a general parser.
Our architecture is similar to the work of Milicevic as both solutions compose a  parser and an inference module.
Our parser is based in NLTK and Spacy while the inference modules uses MACE4 to find the solution and Prover9 to graphically represent the proof for the solution. 

Bogaerts et al. have proposed a solver for logic grid puzzles which also makes use of QA and XAI~\cite{bogaerts2020framework}. 
De Cat et al. employs an extension of FOL~\cite{de2018predicate}. 
A drawback that the named entities (e.g. persons, colors) 
in the puzzle cannot be detected automatically and must be stated by the user.
Also, there exists a semi-automated process to detect the synonymy between verbs.
Differently, we empower our agent with capabilities to automatically detect synonymy between other part-of-speech too, such as nouns. 

Jabrayilzade and Tekir have proposed the  DistilBERT tool that automatically solves logic grid  puzzles~\cite{jabrayilzade2020lgpsolver}. 
The clues are translated in Prolog and the reported accuracy is 100\%.
The zebra puzzles have different categories that need to be recognised within the text (e.g. person, name, occupation, color). 
One assumption of Jabrayilzade and Tekir is that these categories and also their instances are clearly given before parsing the text. 
Differently, we relaxed this assumption by performing named entity recognition. 
This task introduces some points of failure (89.52\% in our experiments), but it represents a mandatory step towards automatic generic solvers. 
A second assumption of Jabrayilzade and Tekir is that the domain is closed to five predicates only: \textit{is}, \textit{either}, \textit{all different}, \textit{pair different}, and \textit{comparison}. 
The solver is based on classifying the clues into one of these five predefined predicates. 
This is performed with a feed-forward neural network with Softmax activation on top. 
A number of 50 puzzles were used for training and 100 for testing. 
In our case, we analysed 43 puzzles to identify the recurrent predicates, and we tested the resulted grammar on 331 puzzles. 

Mitra and Baral have developed the Logicia system~\cite{mitra2015learning} also for 150 Zebra puzzles. 
The clues are classified using a maximum entropy model based on features like POS tags or dependency trees. 
The target language is answer set programming, based on which  71 out of 100 puzzles have been solved.

Some work has been made on the particular puzzles of "knights and knaves".
For instance, Chesani et al. have proposed the following workflow: 
i) understanding the text,
ii) identifying modeling (e.g. proposition logic) and solving techniques (e.g. truth tables, resolution), 
iii) identifying problem components and hidden knowledge (if exists), 
iv) framing the model and solving the problem~\cite{chesani2017solving}. 
A completely different approach, via Logic Algebra, is the work of Ciraulo and Maschio~\cite{ciraulo2020solving}. 
This method relies on  a system of equation for solving a puzzle. 
Thus, encoding every inhabitant from puzzle with a different unknown variable, assuming that a truth proposition is equal to 1, using the properties of a boolean ring and a binary encoding for knight and knave, the solution is found by solving the equations. The disadvantage of this proposal is that it cannot solve other types of logical word puzzles, being focused just on the types and variations of puzzles with knights and knaves.

Our system is fairly successful on the benchmarks of 382 problems,
but these benchmarks are limited to a single category of puzzles. In
particular, "knights and knaves" domain require very little
background knowledge: the main axioms are common to all problems and
the only additional knowledge used is the synonymy of some words. 
We are interested to apply the method to other domains, by automatically import background knowledge.

Our approach belongs to the larger domains of Question Answering (QA) and Explainable AI (XAI). 
QA highlights the ability of a system to answer the questions addressed by humans in natural language, while XAI explains how the solution was found in a human understandable way.
Regarding QA, our software agent is able to provide Yes/No answers to natural language questions related to each puzzle. 
Moreover, in line with XAI, the agent can back its answer, providing a graphical representation of the proof. 

\section{Conclusion}
We described here a tool to automatically solve logical puzzles, in particular knights and knaves puzzles. 
The puzzles are given in natural language. The problem is parsed and translated to first-order logic using the
Natural Language Toolkit with a hand-crafted lexicon and grammar. The MACE4 model finder constructs a solution. 
In addition, the system allows to process natural language yes/no questions by translating them to first-order logic and letting the theorem prover Prover9 find the answer. 
By combining natural language processing with theorem proving, the system can fully explain its answer in the form of a graphical proof. 
The evaluation experiments shown that 309 out of 382 problems could be solved, with only 43 of them were used to build the grammar, and the rest were completely new to the system.

Our solution is based on manually created grammar rules for a closed domain: knights and knaves puzzles. 
The lexicon benefits from the Wordnet lexical database, while the natural language processing pipeline, including coreference resolution, was developed with the Spacy library. 
The resulting FOL theory is given to MACE4 model finder. 
Since MACE4 works on finite domain, it is important to compute the domain size of each problem. 
This is in our case, the exact number of persons that need to be instantiated as knights or knaves. 
Hence, the named entity recognition was a necessary task to automatically compute the domain size. 
We also empower our software agent with the capability  to provide Yes/No answers to natural language questions related to each puzzle. 
Moreover, in line with XAI, the agent can back its answer, providing a graphical representation of the proof. 
These features together with the overall performance rate of 80.89\% make the proposed solution an improvement upon similar solvers for natural language understanding in the puzzles domain.


\section*{Acknowledgments}
We thank the anonymous reviewers for their valuable comments.

\bibliographystyle{IEEEtran}
\bibliography{bib} 

\end{document}